\documentclass[11pt,a4paper]{article}
\usepackage[hyperref]{emnlp-ijcnlp-2019}
\usepackage{times}
\usepackage{latexsym}
\usepackage{graphicx}
\usepackage{amsmath}
\usepackage{amssymb}
\usepackage{bm}
\usepackage{diagbox}

\usepackage{subcaption}

\usepackage{multirow}
\usepackage{url}

\usepackage{paralist}

\usepackage{url}

\DeclareMathOperator\step{\operatorname{step}}
\DeclareMathOperator\warmup{\operatorname{warmup}}

\aclfinalcopy 

\title{Text Summarization with Pretrained Encoders}

\author{Yang Liu\and Mirella Lapata \\
    Institute for Language, Cognition and Computation
    \\
    School of Informatics, University of Edinburgh
    \\ 
    \texttt{yang.liu2@ed.ac.uk, mlap@inf.ed.ac.uk}
}

\date{}

\makeatletter
\newcommand{\thickhline}{%
    \noalign {\ifnum 0=`}\fi \hrule height 1pt
    \futurelet \reserved@a \@xhline
}
\makeatother

\begin{document}
    \maketitle
    \begin{abstract}

        Bidirectional Encoder Representations from Transformers
        (\textsc{Bert}; \citealt{devlin2018bert}) represents the latest
        incarnation of pretrained language models which have recently
        advanced a wide range of natural language processing tasks.  In
        this paper, we showcase how \textsc{Bert} can be usefully applied
        in text summarization and propose a general framework for both
        extractive and abstractive models.  We introduce a novel
        document-level encoder based on \textsc{Bert} which is able to
        express the semantics of a document and obtain representations
        for its sentences. Our extractive model is built on top of this
        encoder by stacking several inter-sentence Transformer layers.
        For abstractive summarization, we propose a new fine-tuning
        schedule which adopts different optimizers for the encoder and
        the decoder as a means of alleviating the mismatch between the
        two (the former is pretrained while the latter is not). We also
        demonstrate that a two-staged fine-tuning approach can further
        boost the quality of the generated summaries. Experiments on
        three datasets show that our model achieves state-of-the-art
        results across the board in both extractive and abstractive
        settings.\footnote{Our code is available at
            \url{https://github.com/nlpyang/PreSumm}.}
    \end{abstract}
    
    \section{Introduction}
    \label{sec:introduction}
    
    Language model pretraining has advanced the state of the art in many
    NLP tasks ranging from sentiment analysis, to question answering,
    natural language inference, named entity recognition, and textual
    similarity. State-of-the-art pretrained models include ELMo
    \cite{peters2018deep}, GPT \cite{radford2018improving}, and more
    recently Bidirectional Encoder Representations from Transformers
    (\textsc{Bert}; \citealt{devlin2018bert}). \textsc{Bert} combines both
    word and sentence representations in a single very large Transformer
    \cite{vaswani2017attention}; it is pretrained on vast amounts of text,
    with an unsupervised objective of masked language modeling and
    next-sentence prediction and can be fine-tuned with various
    task-specific objectives.
    
    In most cases, pretrained language models have been employed as
    encoders for sentence- and paragraph-level natural language
    understanding problems \cite{devlin2018bert} involving various
    classification tasks (e.g.,~predicting whether any two sentences are
    in an entailment relationship; or determining the completion of a
    sentence among four alternative sentences).  
    In this paper, we examine
    the influence of language model pretraining on text
    summarization. Different from previous tasks, summarization requires
    wide-coverage natural language understanding going beyond the meaning
    of individual words and sentences. 
    The aim is to condense a
    \emph{document} into a shorter version while preserving most of its
    meaning. Furthermore, under abstractive modeling formulations, the task
    requires language generation capabilities in order to create summaries
    containing novel words and phrases not featured in the source text, while extractive summarization is often defined as a binary
    classification task with labels indicating whether a text span
    (typically a sentence) should be included in the summary.

    We explore the potential of \textsc{Bert} for text summarization under
    a general framework encompassing both extractive and abstractive
    modeling paradigms.  We propose a novel document-level encoder based
    on \textsc{Bert} which is able to encode a document and obtain
    representations for its sentences. Our extractive model is built on
    top of this encoder by stacking several inter-sentence Transformer
    layers to capture document-level features for extracting sentences.
    Our abstractive model adopts an encoder-decoder architecture,
    combining the same pretrained \textsc{Bert} encoder with a
    randomly-initialized Transformer decoder
    \cite{vaswani2017attention}. We design a new training schedule which
    separates the optimizers of the encoder and the decoder in order to
    accommodate the fact that the former is pretrained while the latter
    must be trained from scratch.  Finally, motivated by previous work
    showing that the combination of extractive and abstractive objectives
    can help generate better summaries~\cite{gehrmann2018bottom}, we
    present a two-stage approach where the encoder is fine-tuned twice,
    first with an extractive objective and subsequently on the abstractive
    summarization task.
    

    We evaluate the proposed approach on three single-document news
    summarization datasets representative of different writing
    conventions (e.g., important information is concentrated at the
    beginning of the document or distributed more evenly throughout)
    and summary styles (e.g., verbose vs. more telegraphic; extractive
    vs. abstractive). Across datasets, we experimentally show that the
    proposed models achieve state-of-the-art results under both
    extractive and abstractive settings. Our contributions in this
    work are three-fold: a)~we highlight the importance of document
    encoding for the summarization task; a variety of recently
    proposed techniques aim to enhance summarization performance via
    copying mechanisms \cite{gu-etal-2016-incorporating,see-acl17,nallapati2017summarunner},
    reinforcement learning \cite{narayan2018ranking,paulus2017deep,dong2018banditsum},
    and multiple communicating encoders \cite{celikyilmaz2018deep}. We
    achieve better results with a minimum-requirement model without
    using any of these mechanisms; b) we showcase ways to effectively
    employ pretrained language models in summarization under both
    extractive and abstractive settings; we would expect any
    improvements in model pretraining to translate in better
    summarization in the future; and c)~the proposed models can be
    used as a stepping stone to further improve summarization
    performance as well as baselines against which new proposals are
    tested.

    %
    %
    %
    %
    %
    %
    
    \begin{figure*}[t]
        \centering
        \includegraphics[width=16cm]{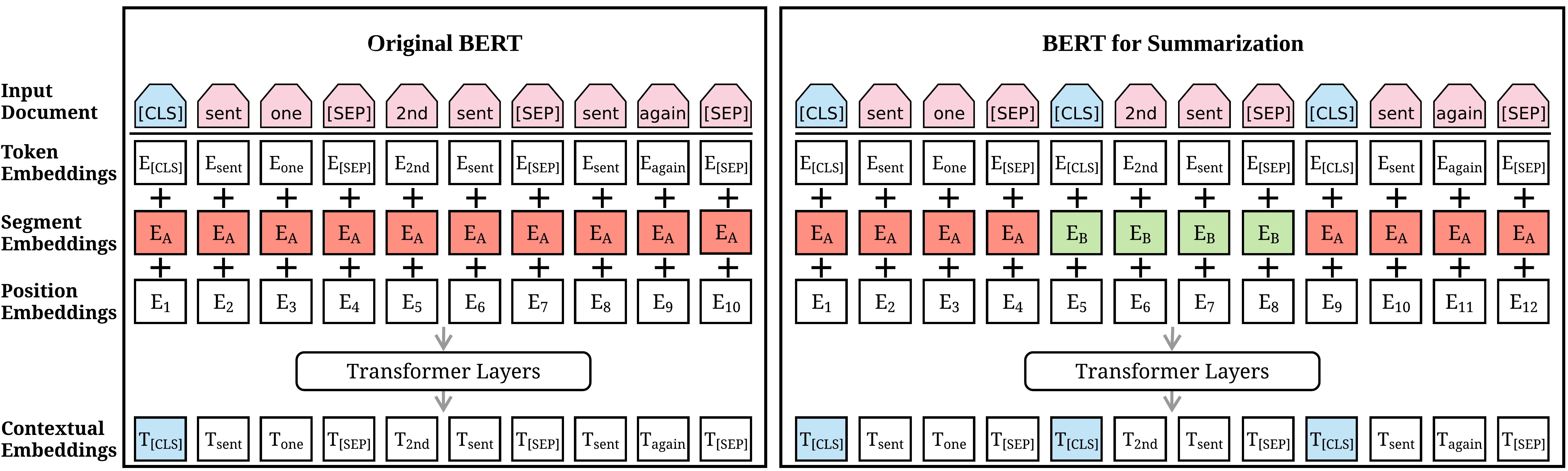}
        \label{trans}
        \caption{\label{fig:bert:architecture} Architecture of the
            original \textsc{Bert} model (left) and \textsc{BertSum}
            (right). The sequence on top is the input document, followed
            by the summation of three kinds of embeddings for each
            token. The summed vectors are used as input embeddings to
            several bidirectional Transformer layers, generating
            contextual vectors for each token. \textsc{BertSum} extends
            \textsc{Bert} by inserting multiple [\textsc{cls}] symbols
            to learn sentence representations and using interval
            segmentation embeddings (illustrated in red and green color)
            to distinguish multiple sentences.}
    \end{figure*}

    \section{Background}
    \label{sec:background}
    
    \subsection{Pretrained Language Models}
    \label{sec:pretr-lang-models}
    Pretrained language models~\cite{peters2018deep,radford2018improving,devlin2018bert,dong2019unified,hibert} have recently emerged as a key technology
    for achieving impressive gains in a wide variety of natural language
    tasks.  These models extend the idea of word embeddings
    by learning contextual representations from large-scale corpora using
    a language modeling objective.  Bidirectional Encoder Representations
    from Transformers (\textsc{Bert}; \citealt{devlin2018bert}) is a new
    language representation model which is trained with a masked language
    modeling  and a ``next sentence prediction'' task on a corpus
    of~3,300M words.
    
    The general architecture of \textsc{Bert} is shown in the left part
    of Figure~\ref{fig:bert:architecture}. Input text is first
    preprocessed by inserting two special tokens.  [\textsc{cls}] is
    appended to the beginning of the text; the output representation of
    this token is used to aggregate information from the whole sequence
    (e.g.,~for classification tasks).  And token~[\textsc{sep}] is
    inserted after each sentence as an indicator of sentence boundaries.
    The modified text is then represented as a sequence of tokens
    $X=[w_1,w_2,\cdots,w_n]$. Each token $w_i$ is assigned three kinds of
    embeddings: \emph{token embeddings} indicate the meaning of each
    token, \emph{segmentation embeddings} are used to discriminate between
    two sentences (e.g.,~during a sentence-pair classification task) and
    \emph{position embeddings} indicate the position of each token within
    the text sequence.  These three embeddings are summed to a single
    input vector $x_i$ and fed to a bidirectional Transformer with
    multiple layers:
            \vspace{-0.5ex}
    \begin{gather}
    \tilde{h}^l=\mathrm{LN}(h^{l-1}+\mathrm{MHAtt}(h^{l-1}))\\
    h^l=\mathrm{LN}(\tilde{h}^l+\mathrm{FFN}(\tilde{h}^l))       
    \end{gather}    
    where $h^0=x$ are the input vectors; $\mathrm{LN}$ is the layer
    normalization operation~\cite{ba2016layer}; $\mathrm{MHAtt}$ is
    the multi-head attention operation~\cite{vaswani2017attention};
    superscript $l$ indicates the depth of the stacked layer. On the
    top layer, \textsc{Bert} will generate an output vector~$t_i$ for
    each token with rich contextual information.
    
    
    Pretrained language models are usually used to enhance performance
in language understanding tasks.  Very recently, there have been
attempts to apply pretrained models to various generation
problems~\cite{edunov2019pre,rothe2019leveraging}.  When
fine-tuning for a specific task, unlike ELMo whose parameters are
usually fixed, parameters in \textsc{Bert} are \emph{jointly}
fine-tuned with additional task-specific parameters.

    \subsection{Extractive Summarization}
    \label{sec:extr-summ-1}
    Extractive summarization systems create a summary by identifying (and
    subsequently concatenating) the most important sentences in a
    document.  
    Neural models consider extractive summarization as a sentence
    classification problem: a neural encoder creates sentence
    representations and a classifier predicts which sentences should be
    selected as summaries.
    \textsc{SummaRuNNer}~\cite{nallapati2017summarunner} is one of the
    earliest neural approaches adopting an encoder based on Recurrent
    Neural Networks.  \textsc{Refresh}~\cite{narayan2018ranking} is a
    reinforcement learning-based system trained by globally optimizing the
    ROUGE metric.  More recent work achieves higher performance with more
    sophisticated model structures.  \textsc{Latent}
    \cite{zhang2018neural} frames extractive summarization as a latent
    variable inference problem; instead of maximizing the likelihood of
    ``gold'' standard labels, their latent model directly maximizes the
    likelihood of human summaries given selected sentences.  
    \textsc{Sumo}
    \cite{yang19sumo} capitalizes on the notion of structured attention to
    induce a multi-root dependency tree representation of the document
    while predicting the output summary.
    \textsc{NeuSum}~\cite{zhou2018neural} scores and selects sentences
    jointly and represents the state of the art in extractive
    summarization.

    \subsection{Abstractive Summarization}
    \label{sec:abstr-summ-1}
    
    Neural approaches to abstractive summarization conceptualize the task
    as a sequence-to-sequence problem,
    where an encoder maps a sequence of tokens in the source document
    $\bm{x} = [x_1, ..., x_n]$ to a sequence of continuous
    representations $\bm{z} = [z_1, ..., z_n]$, and a decoder then
    generates the target summary $\bm{y} = [y_1, ..., y_m]$
    token-by-token, in an auto-regressive manner, hence modeling the
    conditional probability: $p(y_1, ..., y_m|x_1, ..., x_n)$.

    \citet{rush2015neural} and \citet{nallapati2016abstractive} were among
    the first to apply the neural encoder-decoder architecture to text
    summarization.  \citet{see-acl17} enhance this model with a
    pointer-generator network (\textsc{PTgen}) which allows it to copy words
    from the source text, and a coverage mechanism (\textsc{Cov}) which
    keeps track of words that have been summarized.  
    \citet{celikyilmaz2018deep} propose an abstractive system where
    multiple agents (encoders) represent the document together with a
    hierarchical attention mechanism (over the agents) for decoding.
    Their Deep Communicating Agents (\textsc{DCA}) model is trained
    end-to-end with reinforcement learning. \citet{paulus2017deep} also
    present a deep reinforced model (\textsc{DRM}) for abstractive
    summarization which handles the coverage problem with an
    intra-attention mechanism where the decoder attends over previously
    generated words. \citet{gehrmann2018bottom} follow a bottom-up
    approach (\textsc{BottomUp}); a content selector first determines
    which phrases in the source document should be part of the summary, and
    a copy mechanism is applied only to preselected phrases during
    decoding.  \citet{xsum} propose an abstractive model which is
    particularly suited to extreme summarization (i.e.,~single sentence
    summaries), based on convolutional neural networks and additionally
    conditioned on topic distributions (\textsc{TConvS2S}).
    
    

    \section{Fine-tuning \textsc{Bert} for Summarization}
    \label{sec:fine-tuning-textscb}
    
    \subsection{Summarization Encoder}
    \label{sec:summ-encod}
    
    Although \textsc{Bert} has been used to fine-tune various NLP tasks,
    its application to summarization is not as straightforward.  Since
    \textsc{Bert} is trained as a masked-language model, the output
    vectors are grounded to tokens instead of sentences, while in
    extractive summarization, most models manipulate sentence-level
    representations.  Although segmentation embeddings represent different
    sentences in \textsc{Bert}, they only apply to sentence-pair inputs,
    while in summarization we must encode and manipulate multi-sentential
    inputs.  Figure~\ref{fig:bert:architecture} illustrates our proposed
    \textsc{Bert} architecture for \textsc{Sum}marization (which we call
    \textsc{BertSum}).


    In order to represent \emph{individual} sentences, we insert
    external [\textsc{cls}] tokens at the start of each sentence, and
    each [\textsc{cls}] symbol collects features for the sentence
    preceding it. We also use \emph{interval segment} embeddings to
    distinguish multiple sentences within a document. For $sent_i$ we
    assign segment embedding $E_A$ or $E_B$ depending on whether~$i$
    is odd or even. For example, for document $[sent_1, sent_2,
    sent_3, sent_4, sent_5]$, we would assign embeddings $[E_A, E_B,
    E_A,E_B, E_A]$. This way, document representations are learned
    hierarchically where lower Transformer layers represent adjacent
    sentences, while higher layers, in combination with
    self-attention, represent multi-sentence discourse.

    Position embeddings in the original \textsc{Bert} model have a maximum
    length of~512; we overcome this limitation by adding more position
    embeddings that are initialized randomly and fine-tuned with other
    parameters in the encoder.

    \subsection{Extractive Summarization}
    \label{sec:extr-summ}
    
    Let $d$~denote a document containing sentences $[sent_1, sent_2,
    \cdots, sent_m]$, where $sent_i$ is the $i$-th sentence in the
    document.  Extractive summarization can be defined as the task of
    assigning a label $y_i \in \{0, 1\}$ to each $sent_i$, indicating
    whether the sentence should be included in the summary. It is assumed
    that summary sentences represent the most important content of the
    document.
    
    With \textsc{BertSum}, vector $t_i$ which is the vector of the $i$-th
    [\textsc{cls}] symbol from the top layer can be used as the
    representation for $sent_i$.  Several inter-sentence Transformer
    layers are then stacked on top of \textsc{Bert} outputs, to capture
    document-level features for extracting summaries:
    \begin{gather}
    \tilde{h}^l=\mathrm{LN}(h^{l-1}+\mathrm{MHAtt}(h^{l-1}))\\
    h^l=\mathrm{LN}(\tilde{h}^l+\mathrm{FFN}(\tilde{h}^l))
    \end{gather}
    where $h^0=\mathrm{PosEmb}(T)$; $T$ denotes the sentence vectors
    output by \textsc{BertSum}, and function~$\mathrm{PosEmb}$ adds
    sinusoid positional embeddings~\cite{vaswani2017attention} to~$T$,
    indicating the position of each sentence.
    
    The final output layer is a sigmoid classifier:
    \begin{equation}
    \hat{y}_i = \sigma(W_oh_i^L+b_o)
    \end{equation}
    where $h^L_i$ is the vector for $sent_i$ from the top layer (the
    $L$-th layer ) of the Transformer. In experiments, we implemented
    Transformers with $L=1, 2, 3$ and found that a Transformer with
    $L=2$ performed best.
    We name this model \textsc{BertSumExt}.
    
    The loss of the model is the binary classification entropy of
    prediction~$\hat{y}_i$ against gold label~$y_i$.  Inter-sentence
    Transformer layers are jointly fine-tuned with \textsc{BertSum}.
    We use the Adam optimizer with $\beta_1=0.9$, and
    $\beta_2=0.999$). Our learning rate schedule
    follows~\cite{vaswani2017attention} with warming-up ($ \warmup=10,000$):
    \begin{equation}
    \nonumber lr = 2e^{-3}\cdot \min{}(\step{}^{-0.5}, \step{} \cdot \warmup{}^{-1.5})
    \end{equation}

    \begin{table*}[!htbp]
        \begin{small}
            \centerline{
                \begin{tabular}{l|l|cc|cc|c}
                    \thickhline
                    \multicolumn{1}{l|}{\multirow{2}{*}{Datasets}} & \multicolumn{1}{c|}{\multirow{2}{*}{\# docs (train/val/test)}} & \multicolumn{2}{c|}{avg. doc length}                   & \multicolumn{2}{c|}{avg. summary length}                   & \multicolumn{1}{c}{\% novel bi-grams} \\
                    \multicolumn{1}{c|}{}                          & \multicolumn{1}{c|}{}                                          & \multicolumn{1}{c}{words} & \multicolumn{1}{c|}{sentences} & \multicolumn{1}{c}{words} & \multicolumn{1}{c|}{sentences} & \multicolumn{1}{c}{in gold summary}   \\ \thickhline
                    CNN                                            & 90,266/1,220/1,093                                             & 760.50                    & 33.98                          & 45.70                     & 3.59                           & 52.90                                \\
                    DailyMail                                      & 196,961/12,148/10,397                                          & 653.33                    & 29.33                          & 54.65                     & 3.86                           & 52.16                                 \\
                    NYT                                            & 96,834/4,000/3,452                                          & 800.04                    & 35.55                          & 45.54                     & 2.44                           & 54.70                                 \\
                    XSum                                           & 204,045/11,332/11,334                                          & 431.07                    & 19.77                          & 23.26                     & 1.00                           & 83.31                                \\\thickhline
                \end{tabular}
            }
        \end{small}
        
        \caption{\label{tab:statistics} Comparison of summarization datasets: size of training, validation,
            and test sets and average 
            document  and summary length (in terms of words and
            sentences). The proportion of novel bi-grams that do not
            appear in source documents but do appear in the gold summaries 
            quantifies corpus bias towards extractive
            methods.}  
    \end{table*}

    \subsection{Abstractive Summarization}
    \label{sec:abstr-summ}
    
    We use a standard encoder-decoder framework for abstractive
    summarization \cite{see-acl17}. The encoder is the pretrained
    \textsc{BertSum} and the decoder is a 6-layered Transformer
    initialized randomly.  It is conceivable that there is a mismatch
    between the encoder and the decoder, since the former is pretrained
    while the latter must be trained from scratch. This can make
    fine-tuning unstable; for example, the encoder might overfit the data
    while the decoder underfits, or vice versa.  To circumvent this, we
    design a new fine-tuning schedule which separates the optimizers of
    the encoder and the decoder.
    
    
    We use two Adam optimizers with $\beta_1=0.9$ and
    $\beta_2=0.999$ for the encoder and the decoder, respectively, each
    with different warmup-steps and learning rates:
    \begin{align}
    &\hspace*{-.4cm}lr_{\mathcal{E}} = \tilde{lr}_{\mathcal{E}}\cdot \min(step^{-0.5}, \step \cdot \warmup_{\mathcal{E}}^{-1.5})\\
    &\hspace*{-.4cm}lr_{\mathcal{D}} = \tilde{lr}_{\mathcal{D}}\cdot \min(step^{-0.5}, \step \cdot \warmup_{\mathcal{D}}^{-1.5})
    \end{align}
    where $\tilde{lr}_{\mathcal{E}}=2e^{-3}$, and
    $\warmup_{\mathcal{E}}=20,000$ for the encoder and
    $\tilde{lr}_{\mathcal{D}}=0.1$, and $\warmup_{\mathcal{D}}=10,000$ for
    the decoder. This is based on the assumption that the pretrained
    encoder should be fine-tuned with a smaller learning rate and smoother
    decay (so that  the encoder can be trained with more accurate gradients when the decoder is becoming stable).
    
    In addition, we propose a two-stage fine-tuning approach, where we
    first fine-tune the encoder on the extractive summarization task
    (Section~\ref{sec:extr-summ}) and then fine-tune it on the abstractive
    summarization task (Section~\ref{sec:abstr-summ}). Previous
    work~\cite{gehrmann2018bottom,li2018improving} suggests that using
    extractive objectives can boost the performance of abstractive
    summarization. Also notice that this two-stage approach is
    conceptually very simple, the model can take advantage of information
    shared between these two tasks, without fundamentally changing its
    architecture.  We name the default abstractive model
    \textsc{BertSumAbs} and the two-stage fine-tuned model
    \textsc{BertSumExtAbs}.

    \section{Experimental Setup}
    In this section, we describe the summarization datasets used in our
    experiments and discuss various implementation details.
    

    \subsection{Summarization Datasets}
    We evaluated our model on three benchmark datasets, namely the
    CNN/DailyMail news highlights dataset \cite{hermann2015teaching},
    the New York Times Annotated Corpus (NYT; \citealt{nytcorpus}),
    and XSum~\cite{xsum}. These datasets represent different summary
    styles ranging from highlights to very brief one sentence
    summaries. The summaries also vary with respect to the type of
    rewriting operations they exemplify (e.g., some showcase more cut
    and paste operations while others are genuinely
    abstractive). Table~\ref{tab:statistics} presents statistics on
    these datasets (test set); example (gold-standard) summaries are
    provided in the supplementary material.

    \paragraph{CNN/DailyMail} contains news articles and associated
    highlights, i.e.,~a few bullet points giving a brief overview of
    the article.  We used the standard splits
    of~\citet{hermann2015teaching} for training, validation, and
    testing (90,266/1,220/1,093 CNN documents and
    196,961/12,148/10,397 DailyMail documents). We did not anonymize
    entities.  We first split sentences with the Stanford CoreNLP
    toolkit \cite{manning-etal-2014-stanford} and pre-processed the
    dataset following \citet{see-acl17}.  Input documents were
    truncated to~512 tokens.

    \vspace{-1ex}
    \paragraph{NYT} contains 110,540 articles with abstractive
    summaries. Following~\citet{durrett2016learning}, we split these
    into 100,834/9,706 training/test examples, based on the date of
    publication (the test set contains all articles published from
    January 1, 2007 onward).  We used 4,000~examples from the training
    as validation set.  We also followed their filtering procedure,
    documents with summaries less than~50 words were removed from the
    dataset.  The filtered test set (NYT50) includes~3,452 examples.
    Sentences were split with the Stanford CoreNLP toolkit
    \cite{manning-etal-2014-stanford} and pre-processed following
    \citet{durrett2016learning}.  Input documents were truncated
    to~800 tokens.
    
    \vspace{-1ex}
    \paragraph{XSum} contains 226,711 news articles accompanied with a
    one-sentence summary, answering the question ``What is this
    article about?''.  We used the splits of~\citet{xsum} for
    training, validation, and testing (204,045/11,332/11,334) and
    followed the pre-processing introduced in their work.  Input
    documents were truncated to~512 tokens.

    Aside from various statistics on the three datasets,
    Table~\ref{tab:statistics} also reports the proportion of novel
    bi-grams in gold summaries as a measure of their
    abstractiveness. We would expect models with extractive biases to
    perform better on datasets with (mostly) extractive summaries, and
    abstractive models to perform more rewrite operations on datasets
    with abstractive summaries. CNN/DailyMail and NYT are somewhat
    extractive, while XSum is highly abstractive.

    \subsection{Implementation Details} 
    For both extractive and abstractive settings, we used PyTorch,
    OpenNMT~\cite{klein2017opennmt} and the
    `bert-base-uncased'\footnote{\url{https://git.io/fhbJQ}} version of
    \textsc{Bert} to implement \textsc{BertSum}.  Both source and
    target texts were tokenized with \textsc{Bert}'s subwords tokenizer.

    \paragraph{Extractive Summarization}
    All extractive models were trained for 50,000 steps on 3 GPUs (GTX
    1080 Ti) with gradient accumulation every two steps.  Model
    checkpoints were saved and evaluated on the validation set every
    1,000 steps. We selected the top-3 checkpoints based on the
    evaluation loss on the validation set, and report the averaged
    results on the test set.    
    We used a greedy
    algorithm similar to \citet{nallapati2017summarunner} to obtain an
    oracle summary for each document to train extractive models. The algorithm generates an oracle
    consisting of multiple sentences which maximize the ROUGE-2 score
    against the gold summary.

    When predicting summaries for a new document, we first use the
    model to obtain the score for each sentence.  We then rank these
    sentences by their scores from highest to lowest, and select the
    top-3 sentences as the summary.
    
    During sentence selection we use \textit{Trigram Blocking} to
    reduce redundancy \cite{paulus2017deep}.  Given summary~$S$ and
    candidate sentence~$c$, we skip~$c$ if there exists a trigram
    overlapping between $c$ and~$S$. The intuition is similar to
    Maximal Marginal Relevance (MMR; \citealt{carbonell1998use}); we
    wish to minimize the similarity between the sentence being
    considered and sentences which have been already selected as part
    of the summary.
    
    \paragraph{Abstractive Summarization}
    In all abstractive models, we applied dropout (with
    probability~$0.1$) before all linear layers; label
    smoothing~\citep{szegedy2016rethinking} with smoothing
    factor~$0.1$ was also used.  Our Transformer decoder has
    768~hidden units and the hidden size for all feed-forward layers
    is~2,048.  All models were trained for 200,000 steps on 4 GPUs
    (GTX 1080 Ti) with gradient accumulation every five steps.  Model
    checkpoints were saved and evaluated on the validation set every
    2,500 steps. We selected the top-3 checkpoints based on their
    evaluation loss on the validation set, and report the averaged
    results on the test set.
    
    During decoding we used beam search (size~5), and tuned the
    $\alpha$ for the length penalty~\citep{wu2016google} between $0.6$
    and 1 on the validation set; we decode until an end-of-sequence
    token is emitted and repeated trigrams are
    blocked~\cite{paulus2017deep}.  It is worth noting that our
    decoder applies neither a copy nor a coverage mechanism
    \cite{see-acl17}, despite their popularity in abstractive
    summarization. This is mainly because we focus on building a
    minimum-requirements model and these mechanisms may introduce
    additional hyper-parameters to tune.  Thanks to the subwords
    tokenizer, we also rarely observe issues with out-of-vocabulary
    words in the output; moreover, trigram-blocking produces diverse
    summaries managing to reduce repetitions.

    \begin{table}[t]
        \center
        \renewcommand{\arraystretch}{1.1}
        \begin{small}
            \begin{tabular}{@{}l@{}c@{~~}c@{~~}c@{}} \thickhline
                \multicolumn{1}{@{}l|}{Model} & R1 & R2 & RL \\
                \thickhline
                \multicolumn{1}{@{}l|}{\textsc{Oracle}}                                  & ~52.59 & 31.24 & 48.87 \\
                \multicolumn{1}{@{}l|}{\textsc{Lead-3}} & ~40.42 & 17.62 &
                36.67 \\ \hline \multicolumn{4}{@{}c}{Extractive} \\\hline
                \multicolumn{1}{@{}l@{}|}{\textsc{SummaRuNNer} \cite{nallapati2017summarunner}}                             & ~39.60 & 16.20 & 35.30 \\
                \multicolumn{1}{@{}l|}{\textsc{Refresh} \cite{narayan2018ranking}}                                 & ~40.00 & 18.20 & 36.60 \\
                \multicolumn{1}{@{}l|}{\textsc{Latent}
                    \cite{zhang2018neural}}
                & ~41.05 & 18.77 & 37.54 \\
                \multicolumn{1}{@{}l|}{\textsc{NeuSum} \cite{zhou2018neural}}                                 & ~41.59& 19.01& 37.98\\
                \multicolumn{1}{@{}l|}{\textsc{Sumo} \cite{yang19sumo}} &~41.00& 18.40 &37.20 \\
                \multicolumn{1}{@{}l|}{Transformer\textsc{Ext}}&
                ~40.90&18.02& 37.17\\ \hline \multicolumn{4}{@{}c}{
                    Abstractive} \\ \hline
                \multicolumn{1}{@{}l|}{\textsc{PTGen} \cite{see-acl17}}    & ~36.44 & 15.66 & 33.42 \\
                \multicolumn{1}{@{}l|}{\textsc{PTGen}+\textsc{Cov} \cite{see-acl17}} & ~39.53 & 17.28 & 36.38 \\
                \multicolumn{1}{@{}l|}{\textsc{DRM} \citep{paulus2017deep}             }             & ~39.87 & 15.82 & 36.90 \\
                \multicolumn{1}{@{}l|}{\textsc{BottomUp} \cite{gehrmann2018bottom} }                & ~41.22 & 18.68 & 38.34 \\
                \multicolumn{1}{@{}l|}{\textsc{DCA}
                    \cite{celikyilmaz2018deep}}                 & ~41.69 & 19.47 & 37.92 \\
                \multicolumn{1}{@{}l|}{Transformer\textsc{Abs}}                 & ~40.21&17.76&37.09\\ \hline
                \multicolumn{4}{@{}c}{\textsc{Bert}-based}                                                       \\ \hline
                \multicolumn{1}{@{}l|}{\textsc{BertSumExt}}                           
                & ~43.25 & 20.24 & 39.63 \\
                \multicolumn{1}{@{}l|}{\textsc{BertSumExt} w/o interval embeddings}                              & ~43.20 & 20.22 & 39.59 \\
                \multicolumn{1}{@{}l|}{\textsc{BertSumExt} (large)}                             & ~43.85 & 20.34 & 39.90 \\
                \multicolumn{1}{@{}l|}{\textsc{BertSumAbs}}                              & ~41.72 & 19.39 & 38.76 \\
                \multicolumn{1}{@{}l|}{\textsc{BertSumExtAbs}}
                & ~42.13 & 19.60  & 39.18 \\ \thickhline 
            \end{tabular}
        \end{small}
        \caption{\label{tab:cnndaily} ROUGE F1 results on  \textbf{CNN/DailyMail} test set (R1 and R2 are shorthands for unigram and bigram
            overlap; RL is the longest common subsequence). Results for
            comparison systems are  taken from the authors' respective papers or
            obtained on our data by 
            running publicly released software.} 
    \end{table}

    \begin{table}[t]
        \center
        \renewcommand{\arraystretch}{1.1}
        \begin{small}
            \begin{tabular}{lccc@{}} \thickhline
                \multicolumn{1}{l|}{Model}                                   & R1    & R2    & RL    \\ \thickhline
                \multicolumn{1}{l|}{\textsc{Oracle}}                                  &49.18& 33.24& 46.02 \\
                \multicolumn{1}{l|}{\textsc{Lead-3}}                                   & 39.58& 20.11 &35.78 \\ \hline
                \multicolumn{4}{c}{Extractive}                                                       \\\hline
                \multicolumn{1}{l|}{\textsc{Compress} \cite{durrett2016learning}}& 42.20  &24.90  &  --- \\ 
                \multicolumn{1}{l|}{\textsc{Sumo} \cite{yang19sumo}} &42.30& 22.70 &38.60 \\ 
                \multicolumn{1}{l|}{Transformer\textsc{Ext} }& 41.95 &22.68&38.51\\ \hline
                \multicolumn{4}{c}{Abstractive}                                                      \\ \hline
                
                \multicolumn{1}{l|}{\textsc{PTGen} \cite{see-acl17}      }             & 42.47 &  25.61 & --- \\
                \multicolumn{1}{l|}{\textsc{PTGen} + \textsc{Cov}  \cite{see-acl17}    }             & 43.71 &  26.40 & --- \\
                \multicolumn{1}{l|}{\textsc{DRM}   \cite{paulus2017deep}
                }             & 42.94 &  26.02 & --- \\
                \multicolumn{1}{l|}{Transformer\textsc{Abs}}& 35.75 &17.23&31.41\\ 
                \hline
                \multicolumn{4}{c}{\textsc{Bert}-based}                                                       \\ \hline
                \multicolumn{1}{l|}{\textsc{BertSumExt} }                             & 46.66 &26.35& 42.62 \\
                \multicolumn{1}{l|}{\textsc{BertSumAbs}}                              & 48.92   &30.84  &45.41\\
                \multicolumn{1}{l|}{\textsc{BertSumExtAbs} }
                
                & 49.02 &31.02& 45.55 \\ \thickhline
            \end{tabular}
        \end{small}
        \caption{\label{tab:nyt} ROUGE Recall results on  \textbf{NYT} test set. Results for comparison systems are
            taken from the authors' respective papers or obtained on our data by running publicly released software. Table
            cells are filled with --- whenever results are not available.} 
    \end{table}

    \section{Results}
    \label{sec:results}
    
    \subsection{Automatic Evaluation}
    
    We evaluated summarization quality automatically using ROUGE~\cite{lin:2004:ACLsummarization}.  We report unigram and bigram overlap (ROUGE-1
    and ROUGE-2) as a means of assessing informativeness and the longest
    common subsequence (ROUGE-L) as a means of assessing
    fluency.

    Table~\ref{tab:cnndaily} summarizes our results on the CNN/DailyMail
    dataset.  The first block in the table includes the results of an
    extractive \textsc{Oracle} system as an upper bound.  We also present the \textsc{Lead-3} baseline
    (which simply selects the first three sentences in a document).
    
    The second block in the table includes various extractive models
    trained on the CNN/DailyMail dataset (see
    Section~\ref{sec:extr-summ-1} for an overview). For comparison to our
    own model, we also implemented a non-pretrained Transformer baseline
    (Transformer\textsc{Ext}) which uses the same architecture as
    \textsc{BertSumExt}, but with fewer parameters. It is randomly initialized
    and only trained on the summarization task. Transformer\textsc{Ext}
    has~6 layers, the hidden size is 512, and the feed-forward filter size
    is~2,048. The model was trained with same settings as
    in~\citet{vaswani2017attention}.
    
    The third block in Table~\ref{tab:cnndaily} highlights the performance
    of several abstractive models on the CNN/DailyMail dataset (see
    Section~\ref{sec:abstr-summ-1} for an overview). We also include an
    abstractive Transformer baseline (Transformer\textsc{Abs}) which has
    the same decoder as our abstractive \textsc{BertSum} models; the encoder
    is a 6-layer Transformer with 768 hidden size and 2,048~feed-forward
    filter size.

    The fourth block reports results with fine-tuned \textsc{Bert} models:
    \textsc{BertSumExt} and its two variants  (one without interval
    embeddings, and one with the large version of \textsc{Bert}),
    \textsc{BertSumAbs}, and \textsc{BertSumExtAbs}.  \textsc{Bert}-based
    models outperform the \textsc{Lead}-3 baseline which is not a
    strawman; on the CNN/DailyMail corpus it is indeed superior to
    several extractive
    \cite{nallapati2017summarunner,narayan2018ranking,zhou2018neural} and
    abstractive models \cite{see-acl17}. \textsc{Bert} models collectively
    outperform all previously proposed extractive and abstractive systems,
    only falling behind the \textsc{Oracle} upper bound. Among
    \textsc{Bert} variants, \textsc{BertSumExt} performs best which is not
    entirely surprising; CNN/DailyMail summaries are somewhat extractive
    and even abstractive models are prone to copying sentences from the
    source document when trained on this dataset \cite{see-acl17}. Perhaps
    unsurprisingly we observe that larger versions of \textsc{Bert} lead
    to performance improvements and that interval embeddings bring only
    slight gains.
    
    
    Table~\ref{tab:nyt} presents results on the NYT dataset.  Following
    the evaluation protocol in~\citet{durrett2016learning}, we use
    limited-length ROUGE Recall, where predicted summaries are truncated
    to the length of the gold summaries.  Again, we report the performance
    of the \textsc{Oracle} upper bound and \textsc{Lead}-3 baseline. The
    second block in the table contains previously proposed extractive
    models as well as our own Transformer baseline. \textsc{Compress}
    \cite{durrett2016learning} is an ILP-based model which combines
    compression and anaphoricity constraints. The third block includes
    abstractive models from the literature, and our Transformer baseline.
    \textsc{Bert}-based models are shown in the fourth block. Again, we
    observe that they outperform previously proposed  approaches. On this dataset, abstractive \textsc{Bert}
    models generally perform better compared to \textsc{BertSumExt},
    almost approaching \textsc{Oracle} performance.
    

    \begin{table}[t]
        \center
        \renewcommand{\arraystretch}{1.1}
        \begin{small}
            \begin{tabular}{@{}ll@{~~}r@{~~}l@{}} \thickhline
                \multicolumn{1}{@{}l|}{Model}                                   & \multicolumn{1}{c}{R1}    & \multicolumn{1}{c}{R2}    & \multicolumn{1}{c}{RL}    \\ \thickhline
                \multicolumn{1}{@{}l|}{\textsc{Oracle}}                                  & 29.79& 8.81 &22.66\\
                \multicolumn{1}{@{}l|}{\textsc{Lead}}                                   &16.30 &1.60 &11.95\\ \hline
                \multicolumn{4}{@{}c}{Abstractive}                                                      \\ \hline
                \multicolumn{1}{@{}l|}{\textsc{PTGen}~\cite{see-acl17}}                 & 29.70& 9.21 &23.24 \\
                \multicolumn{1}{@{}l|}{\textsc{PTGen}+\textsc{Cov}~\cite{see-acl17}}                 & 28.10& 8.02 &21.72 \\
                \multicolumn{1}{@{}l|}{\textsc{TConvS2S}~\cite{xsum}}                 & 31.89&
                11.54& 25.75\\
                \multicolumn{1}{@{}l|}{Transformer\textsc{Abs}} &29.41&9.77&23.01 \\
                \hline
                \multicolumn{4}{@{}c}{\textsc{Bert}-based}                                                       \\ \hline
                \multicolumn{1}{@{}l|}{\textsc{BertSumAbs}   }                           & 38.76    &16.33& 31.15\\
                \multicolumn{1}{@{}l|}{\textsc{BertSumExtAbs}  }  &38.81&16.50&31.27\\ \thickhline
            \end{tabular}
        \end{small}
        \caption{\label{tab:xsumresults} ROUGE F1 results on the \textbf{XSum} test set. Results for comparison systems are
            taken from the authors' respective papers or obtained on our data by
            running publicly released software. } 
    \end{table}

    Table~\ref{tab:xsumresults} summarizes our results on the XSum
    dataset. Recall that summaries in this dataset are highly
    abstractive (see Table~\ref{tab:statistics}) consisting of a
    single sentence conveying the gist of the document. Extractive
    models here perform poorly as corroborated by the low performance
    of the \textsc{Lead} baseline (which simply selects the leading
    sentence from the document), and the \textsc{Oracle} (which
    selects a single-best sentence in each document) in
    Table~\ref{tab:xsumresults}. As a result, we do not report results
    for extractive models on this dataset.
    The second block in Table~\ref{tab:xsumresults} presents the results
    of various abstractive models taken from \citet{xsum} and also
    includes our own abstractive Transformer baseline. In the third block
    we show the results of our \textsc{Bert} summarizers which again are
    superior to all previously reported models (by a wide margin).
    
    \begin{table}[t]
        \begin{small}
            \centerline{
                \begin{tabular}{l|llll}
                    \diagbox{$\tilde{lr}_{\mathcal{E}}$}{$\tilde{lr}_{\mathcal{D}}$} & 1     & 0.1  & 0.01 & 0.001 \\ \thickhline
                    2e-2 & 50.69  &9.33 &  10.13 & 19.26 \\
                    2e-3 & 37.21 & 8.73 & 9.52 & 16.88 \\\thickhline
                \end{tabular}
            }
        \end{small}
        \caption{\label{exp-lr} Model perplexity  (CNN/DailyMail; validation set)
            under different combinations of encoder  and decoder
            learning rates.}
    \end{table}

    \subsection{Model Analysis}
    
    \paragraph{Learning Rates} Recall that our abstractive model uses
    separate optimizers for the encoder and decoder. In Table~\ref{exp-lr}
    we examine whether the combination of different learning rates
    ($\tilde{lr}_{\mathcal{E}}$ and $\tilde{lr}_{\mathcal{D}}$) is indeed
    beneficial.  Specifically, we report model perplexity on the
    CNN/DailyMail validation set for varying encoder/decoder learning
    rates.  We can see that the model performs best with
    $\tilde{lr}_{\mathcal{E}}=2e-3$ and $\tilde{lr}_{\mathcal{D}}=0.1$.

\begin{figure}[t]
        \centering
        \hspace*{-1ex}\includegraphics[width=8cm]{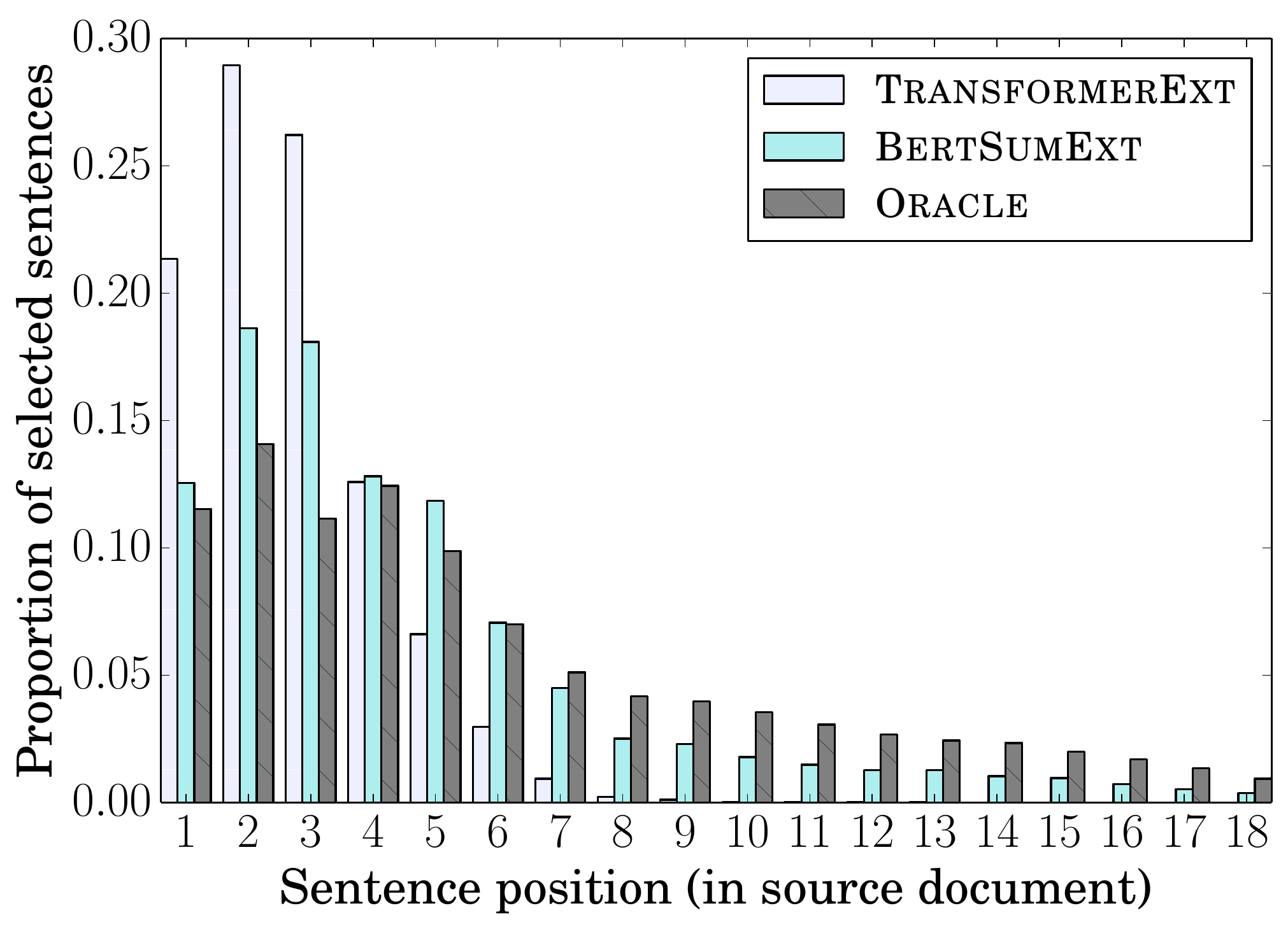}
        \caption{Proportion of extracted sentences according to their
            position in the original document.}
        \label{fig:figure1}
    \end{figure}

    \paragraph{Position of Extracted Sentences}    
    In addition to the evaluation based on ROUGE, we also analyzed in more
    detail the summaries produced by our model. For the extractive
    setting, we looked at the position (in the source document) of the
    sentences which were selected to appear in the
    summary. Figure~\ref{fig:figure1} shows the proportion of selected summary
    sentences which appear in the source document at positions 1, 2, and so on. The
    analysis was conducted on the CNN/DailyMail dataset for Oracle
    summaries, and those produced by \textsc{BertSumExt} and the
    Transformer{\sc Ext}.  We can see that Oracle summary
    sentences are fairly smoothly distributed across documents, while
    summaries created by Transformer\textsc{Ext} mostly concentrate on the first
    document sentences. \textsc{BertSumExt} outputs are more similar to Oracle
    summaries, indicating that with the pretrained encoder, the model
    relies less on shallow position features, and learns deeper document
    representations.
    
\begin{figure}[t]
        \centering
        \begin{subfigure}{0.5\textwidth}
            \centering
            \includegraphics[width=7.5cm]{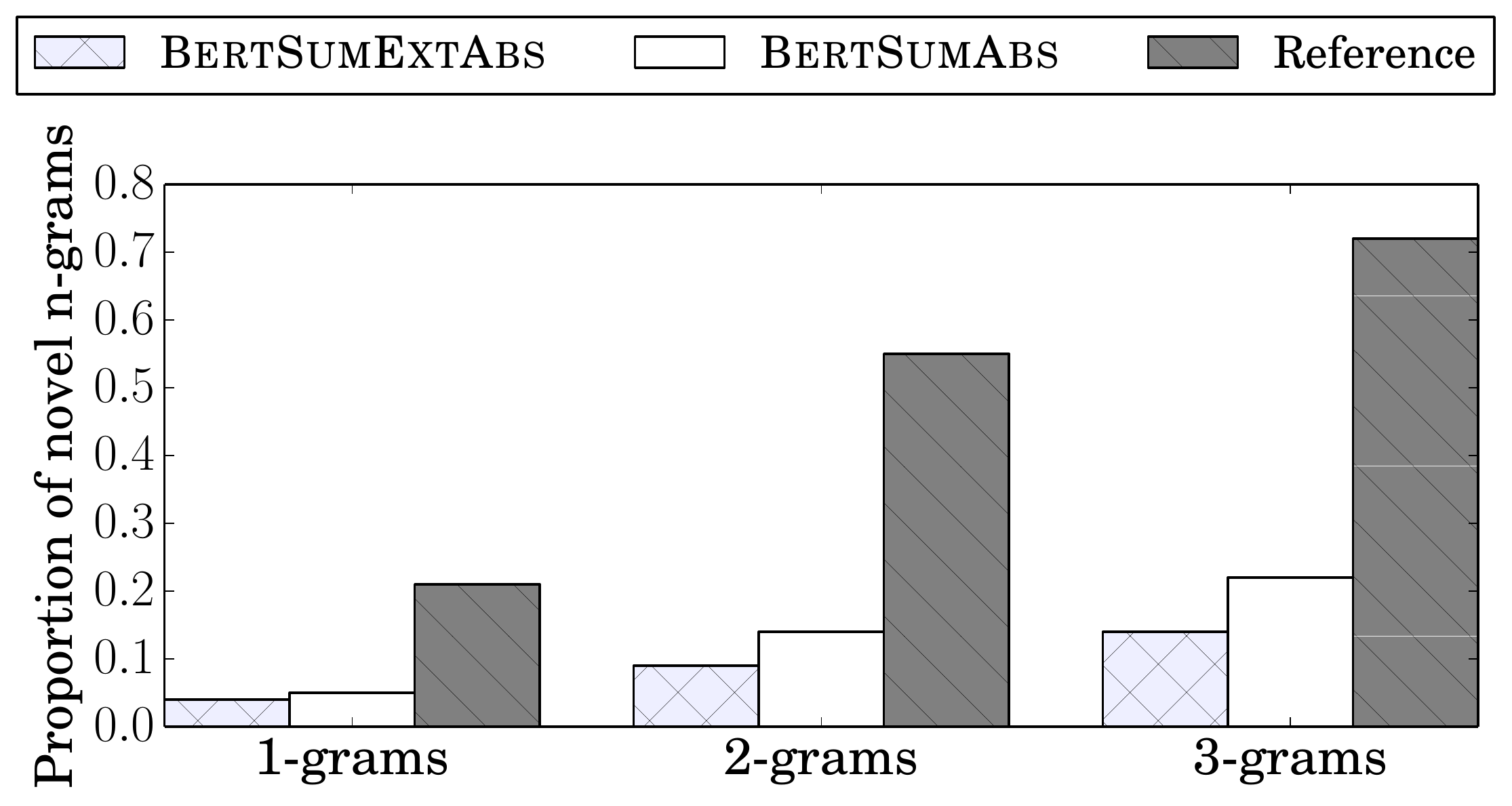}
            \caption{CNN/DailyMail Dataset}
            \label{fig:inclu}
        \end{subfigure}%
        \\
        \begin{subfigure}{0.5\textwidth}
            \centering
            \includegraphics[width=7.3cm]{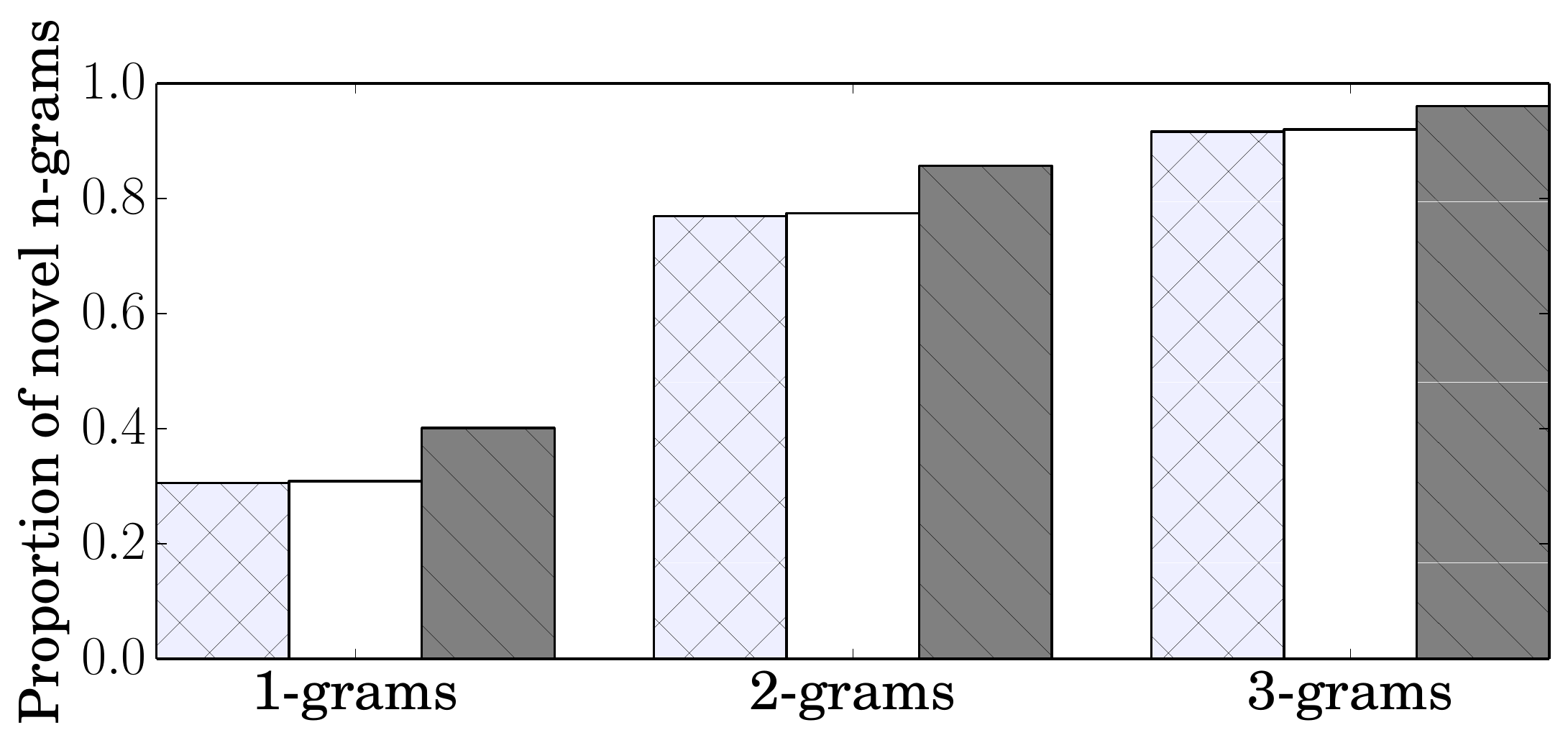}
            \caption{XSum dataset}
            
            \label{fig:deform}
        \end{subfigure}
        \caption{Proportion of novel n-grams in model generated
            summaries.}
        \label{fig:manmade}
    \end{figure}

    \paragraph{Novel N-grams} We also analyzed the output of
    abstractive systems by calculating the proportion of novel n-grams that
    appear in the summaries but not in the source texts.  The results
    are shown in Figure~\ref{fig:manmade}. In the CNN/DailyMail
    dataset, the proportion of novel n-grams in automatically
    generated summaries is much lower compared to reference summaries,
    but in XSum, this gap is much smaller.  We also observe that on
    CNN/DailyMail, \textsc{BertExtAbs} produces less novel n-ngrams
    than \textsc{BertAbs}, which is not surprising. \textsc{BertExtAbs}
    is more biased towards selecting sentences from the source
    document since it is initially trained as an extractive model.
    
    The supplementary material includes examples of system output and
additional ablation studies.

    \subsection{Human Evaluation}   
    \label{sec:human-evaluation}
    
    In addition to automatic evaluation, we also evaluated system output
    by eliciting human judgments.  We report experiments following a
    question-answering (QA)
    paradigm~\citep{clarke2010discourse,narayan2018ranking} which
    quantifies the degree to which summarization models retain key
    information from the document. Under this paradigm, a set of questions
    is created based on the gold summary under the assumption that it
    highlights the most important document content. Participants are then
    asked to answer these questions by reading system summaries alone
    without access to the article. The more questions a system can answer,
    the better it is at summarizing the document as a whole. 
    
    Moreover, we also assessed the overall
    quality of the summaries produced by abstractive systems which due to
    their ability to rewrite content may produce disfluent or
    ungrammatical output.  Specifically, we followed the Best-Worst
    Scaling~\cite{bestworstscaling} method where participants were
    presented with the output of two systems (and the original document)
    and asked to decide which one was better according to the criteria of
    \emph{Informativeness}, \emph{Fluency}, and \emph{Succinctness}.

    \begin{table}[t]
        \begin{small}
            \centerline{
                \begin{tabular}{l|cc}
                    \thickhline
                    Extractive            & \multicolumn{1}{c}{CNN/DM} & \multicolumn{1}{c}{NYT}  \\ \thickhline
                    \textsc{Lead}        & 42.5$^{\dagger}$  & 36.2$^{\dagger}$ \\
                    \textsc{NeuSum}& 42.2$^{\dagger}$    & ---    \\
                    \textsc{Sumo}     & 41.7$^{\dagger}$       & 38.1$^{\dagger}$  \\
                    Transformer & 37.8$^{\dagger}$    & 32.5$^{\dagger}$  \\
                    \textsc{BertSum}     & \hspace*{-.8ex}58.9   & \hspace*{-.8ex}41.9 \\ \hline
                \end{tabular}
            }
        \end{small}
        \caption{QA-based evaluation. Models with~${\dagger}$  are
            significantly different  from \textsc{BertSum} (using a paired
            student t-test; \mbox{$p < 0.05$}). Table cells are filled
            with --- whenever system output is not available.}
        \label{human-ext}
    \end{table}

    Both types of evaluation were conducted on the Amazon Mechanical
    Turk platform. For the CNN/DailyMail and NYT datasets we used the
    same documents (20 in total) and questions from previous work
    \cite{narayan2018ranking,yang19sumo}. For XSum, we randomly
    selected~20 documents (and their questions) from the release
    of~\citet{xsum}. We elicited 3 responses per HIT.  With regard to
    QA evaluation, we adopted the scoring mechanism from
    \citet{clarke2010discourse}; correct answers were marked with a
    score of one, partially correct answers with~0.5, and zero
    otherwise. For quality-based evaluation, the rating of each system
    was computed as the percentage of times it was chosen as better
    minus the times it was selected as worse. Ratings thus range from
    -1 (worst) to 1 (best).
    
    \begin{table}[t]
        \begin{small}
            \centerline{
                \begin{tabular}{@{}l@{}|c@{~~}c|c@{~~}c|c@{~~}c@{}}
                    \thickhline
                    & \multicolumn{2}{c|}{CNN/DM} & \multicolumn{2}{c|}{NYT} & \multicolumn{2}{c}{XSum} \\ 
                    \multicolumn{1}{c|}{Abstractive} & QA          & Rank       & QA       & Rank       & QA         & Rank     \\\thickhline
                    \textsc{Lead}              & 42.5$^{\dagger}$             &  ---             &     36.2$^{\dagger}$      &    ---           & 9.20$^{\dagger}$         &  ---           \\
                    \textsc{PTGen} & 33.3$^{\dagger}$        & -0.24$^{\dagger}$         &  30.5$^{\dagger}$      &    -0.27$^{\dagger}$            & 23.7$^{\dagger}$       & -0.36$^{\dagger}$       \\
                    \textsc{BottomUp}          & 40.6$^{\dagger}$        & -0.16$^{\dagger}$         &   ---   &     ---        & ---      & ---           \\
                    \textsc{TConvS2S}         &   ---     &---          &  ---      &      ---       & \hspace*{-.8ex}52.1      & \hspace*{1ex}-0.20$^{\dagger}$         \\
                    \textsc{Gold}              &     ---   & 0.22$^{\dagger}$    &  ---
                    &      \hspace*{1ex}0.33$^{\dagger}$          &    ---     & \hspace*{1ex}0.38$^{\dagger}$        \\ 
                    \textsc{BertSum}           & \hspace*{-.8ex}56.1       &  \hspace*{-.9ex}0.17        &   41.8       &     \hspace*{-.5ex}-0.07         & \hspace*{-.8ex}57.5       & 0.19       \\ \thickhline
                \end{tabular}
            }
        \end{small}
        \caption{\label{human-abs} QA-based and ranking-based evaluation. Models with~${\dagger}$  are significantly different  from \textsc{BertSum} (using
            a paired student t-test; \mbox{$p < 0.05$}). Table cells are filled
            with --- whenever system output is not available. \textsc{Gold} is
            not used in QA setting, and \textsc{Lead} is not used in Rank evaluation.}
    \end{table}

    Results for extractive and abstractive systems are shown in
    Tables~\ref{human-ext} and~\ref{human-abs}, respectively.  We compared
    the best performing \textsc{BertSum} model in each setting (extractive
    or abstractive) against various state-of-the-art systems (whose output
    is publicly available), the \textsc{Lead} baseline, and the
    \textsc{Gold} standard as an upper bound. As shown in both tables
    participants overwhelmingly prefer the output of our model against
    comparison systems across datasets and evaluation paradigms. All
    differences between \textsc{BertSum} and comparison models are
    statistically significant (\mbox{$p<0.05$}), with the exception of
    \textsc{TConvS2S} (see Table~\ref{human-abs}; XSum) in the QA
    evaluation setting.

    \section{Conclusions}
    \label{sec:conclusions}

    In this paper, we showcased how pretrained \textsc{Bert} can be
    usefully applied in text summarization. We introduced a novel
    document-level encoder and proposed a general framework for both
    abstractive and extractive summarization. Experimental results across
    three datasets show that our model achieves state-of-the-art results
    across the board under automatic and human-based evaluation
    protocols. Although we mainly focused on document encoding for
    summarization, in the future, we would like to take advantage the
    capabilities of \textsc{Bert} for language generation.
    


    
      \section*{Acknowledgments} This research is supported by a Google
    PhD Fellowship to the first author.  We gratefully acknowledge the
    support of the European Research Council (Lapata, award number
    681760, ``Translating Multiple Modalities into Text'').  We would
    also like to thank Shashi Narayan for providing us with the XSum dataset.
    
    \newpage
    \bibliography{references}
    \bibliographystyle{acl_natbib}

\end{document}